\documentclass[11pt,a4paper]{article}
\usepackage[hyperref]{style/naaclhlt2019}
\usepackage{times}
\usepackage{latexsym}

\usepackage{url}
\usepackage{helvet}
\usepackage{courier}
\usepackage{url}
\usepackage{graphicx}
\usepackage[english]{babel}
\usepackage{amsmath}
\usepackage{amssymb}
\usepackage{mathtools}
\usepackage{color}
\usepackage{multirow}
\usepackage[inline]{enumitem}

\newcommand{\textunderscript}[1]{$_{\text{#1}}$}

\newcommand{\ETE}[0]{End-to-end}
\newcommand{\ete}[0]{end-to-end}
\newcommand{\DEID}[0]{de-ID}
\newcommand{\NER}[0]{NER}
\newcommand{\WER}[0]{WER}
\newcommand{\ASR}[0]{ASR}
\newcommand{\PHI}[0]{PHI}

\newcommand{\Audio}[1]{A(#1)}
\newcommand{\word}[1]{w_{#1}}

\newcommand{\IFOURTEEN}[0]{\textbf{\emph{I2B2'14}}}
\newcommand{\CALLISTO}[0]{\textbf{\emph{AMC'17}}}

\newcommand{\SWITCHBOARD}[0]{\textbf{\emph{Switchboard}}}
\newcommand{\FISHER}[0]{\textbf{\emph{Fisher}}}
\newcommand{\SANDF}[0]{\textbf{\emph{SWFI}}}
\newcommand{\ESTER}[0]{\textbf{\emph{ESTER}}}
\newcommand{\ETAPE}[0]{\textbf{\emph{ETAPE}}}

\newcommand{\FULLCOVERED}[1]{covered(#1)}
\newcommand{\PARTCOVERED}[1]{\rho\!-\!covered(#1)}
\newcommand{\IDENT}[0]{PHI}
\newcommand{\IDENTS}[0]{PHIs}
\newcommand{\NONIDENT}[0]{non-PHI}
\newcommand{\RECALL}[0]{$Recall$}
\newcommand{\RECALLRHO}[0]{$Recall_{\rho}$}
\newcommand{\PRECISION}[0]{$Precision$}
\newcommand{\PRECISIONRHO}[0]{$Precision_{\rho}$}
\newcommand{\FOne}[0]{$F1$}

\newcommand{\CallistoModel}[0]{\textbf{\emph{M\textunderscript{AMC}}}}
\newcommand{\ItbtModel}[0]{\textbf{\emph{M\textunderscript{I2B2}}}}
\newcommand{\ConversationModel}[0]{\textbf{\emph{M\textunderscript{SWFI}}}}
\newcommand{\ConversationModelOrig}[0]{\textbf{\emph{M\textunderscript{SWFIReg}}}}
\newcommand{\ConversationModelMix}[0]{\textbf{\emph{M\textunderscript{SWFI\_MixCase}}}}
\newcommand{\ConversationModelMixAsr}[0]{\textbf{\emph{M\textunderscript{SWFI\_MixCase+Asr}}}}

\newcommand{\DataPublicUrl}{\scriptsize{\url{https://g.co/audio-ner-annotations-data}}}

\newcommand\ignore[1]{}

\title{Audio De-identification: A New Entity Recognition Task}
\author{
Ido Cohn, Itay Laish, Genady Beryozkin, Gang Li, Izhak Shafran, \\
{\bf Idan Szpektor}, {\bf Tzvika Hartman}, {\bf Avinatan Hassidim}, {\bf Yossi Matias} \\
  Google \\
  Tel Aviv, Israel \\
  {\tt \{idoc,itaylaish,leebird,tzvika\}@google.com}
}
\date{}

\begin{document}
\maketitle
\begin{abstract}
Named Entity Recognition (\NER{}) has been mostly studied in the context of written text. Specifically, \NER{} is an important step in de-identification (\DEID{}) of medical records, many of which are recorded conversations between a patient and a doctor. In such recordings, audio spans with personal information should be redacted, similar to the redaction of sensitive character spans in \DEID{} for written text.
The application of \NER{} in the context of audio de-identification has yet to be fully investigated. 
To this end, we define the task of audio \DEID{}, in which audio spans with entity mentions should be detected. We then present our pipeline for this task, which involves Automatic Speech Recognition (\ASR{}), \NER{} on the transcript text, and text-to-audio alignment. Finally, we introduce a novel metric for audio \DEID{} and a new evaluation benchmark consisting of a large labeled segment of the \SWITCHBOARD{} and \FISHER{} audio datasets and detail our pipeline's results on it.

%% Ronit's abstract:
% NER for the de-identification (\DEID{}) of text medical records has been well studied. Many medical records, however, are recorded conversations between a patient and a doctor. In such recordings, audio spans with personal information should be redacted, similar to the redaction of sensitive character spans in \DEID{} for written text. Performance of NER in the context of a complete audio \DEID{} pipeline has not been well characterised. To this end, we define the task of audio \DEID{}, in which audio spans with entity mentions should be detected. We define a new metric for measuring performance on the audio DEID task. We then present our pipeline for this task, which involves Automatic Speech Recognition (\ASR{}), \NER{} on the transcript text, and text-to-audio alignment. Finally, we introduce a new evaluation benchmark consisting of a large labeled segment of the \SWITCHBOARD{} and \FISHER{} audio datasets and detail our pipeline's results on it

\end{abstract}

\section{Introduction}
\label{sec:intro}

% \begin{enumerate}
%     \item Text deid is known important, today performed on text medical records with high performance
%     \item Real world situation/application - audio records exist (how much? standard?) and have lots of data, so we want to do it on audio, not done yet, deid means redaction of audio stream to remove phi.
%     \item This is a new task without metrics or benchmarks. 
%     \item We built an English test benchmark (SW,Fi), labeled NER, and defined a new evaluation metric on it.
%     \item we chose a pipeline approach for this task (+description) - reuse ASR technologies, visibility. So we can use the abundance of text, and use current ASR for transcription and novel component of alignment back to audio.
%     \item Experimented and have initial results, do not reach text NER level due to the complex stack. We thoroughly analyzed the issues, etc.
% \end{enumerate}

Personal data in general, and clinical records data in particular, is a major driving force in today's scientific research. Despite its  abundance, the presence of Personal Health Identifiers (\PHI{}) hinders data availability for researchers. Therefore, data de-identification (\DEID{}) is a critical component in any plan to make such data available. 
However, the amount of data involved makes it prohibitively expensive to employ domain experts to tag and redact \PHI{} manually, providing a good opportunity for automatic de-identification tools. Indeed, high performance tools for the de-identification of medical text notes have been developed \cite{Dernoncourt2017, Liu2017deid}. 

Due to the rise of tele-medicine \cite{weinstein2014telemedicine}, clinical records consist of many other types of data, such as audio conversations \cite{chiu2017speech}, scanned documents, video, and images. 
In this work, we direct our attention towards the task of de-identifying clinical audio data. 
This task is expected to become increasingly more important, as 
Machine Learning applications in tele-medicine are growing in popularity. 
%and as one considers the possibility of passing a ``medical Turing Test'' of replacing the physician in the clinic with a computer program: 
% Training data for such systems could include recorded conversations that occur in clinics, thus the need to de-identify such conversations arises. 
Given an input audio stream, the objective is to produce a modified audio stream, where all \PHI{} is redacted, while the rest of the stream is kept unchanged. 
To the best of our knowledge, de-identifying audio is a new task, requiring a new benchmark.
% It is similar to the Audio \NER{} task, that involves recognizing entities in audio transcripts, with the important distinction that the output is not tagged tokens, but an actual redacted audio stream. 

We define and publish\footnote{\DataPublicUrl{}\label{ftn:data-public-url}} a benchmark consisting of the following:
\begin{enumerate*}
    \item A large labeled subset of the \SWITCHBOARD{} \cite{Godfrey:1992:STS:1895550.1895693} and \FISHER{} \cite{David04thefisher} conversational English audio datasets, denoted as \SANDF{}.
    \item A new evaluation metric, measuring how well the \IDENT{} words in the input audio were identified and redacted, and how well the rest of the audio was preserved.
\end{enumerate*} 

% The Callisto dataset [ref] consisting of recorded patient-doctor conversations.

\begin{figure}[t]
    \centering
    % Generated from https://docs.google.com/document/d/15ElHF0wmC7XYRi2PN7ZYvQbpI3U3Xt2FrTER5Zaeo-0/edit?usp=sharing
    \includegraphics[width=0.43\textwidth]{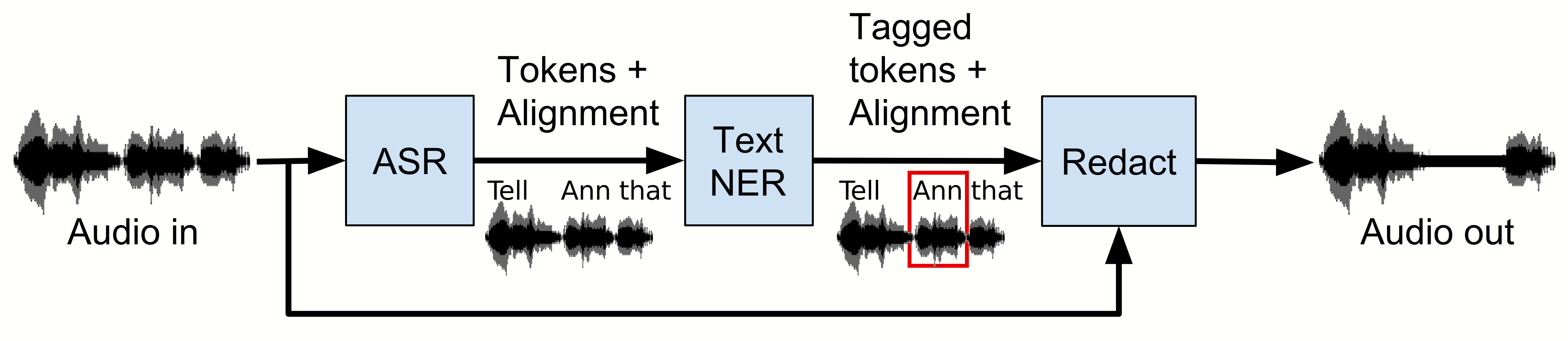}
    \caption{High level audio \DEID{} pipeline}
    \label{fig:pipeline-high-level}
\end{figure}

To better understand the challenges of the audio de-id task, we evaluate it both \ete{} and by breaking it down and solving it using individual components.
Our pipeline (Fig. \ref{fig:pipeline-high-level}) first produces transcripts from the audio using \ASR{}, proceeds by running text-based \NER{} tagging, and then redacts PHI tokens, using the aligned token boundaries determined by \ASR{}.
% , each optimized for its subtask.
Our tagger relies on the state-of-the-art techniques for solving the audio \NER{} problem of recognizing entities in audio transcripts \cite{Lample2016NNbasedNER,Ma2016}. 
We leverage the available Automatic Speech Recognition (\ASR{}) technology, and use its component of alignment back to audio.\\

Finally, we evaluate our pipeline and describe its performance, both \ete{} and per-component. 
Although results on audio are worse than \NER{} performance on text, the pipeline achieves better results than expected despite the compounding pipeline errors. 
Last, we analyze our performance and provide insights for next steps.

\section{Related Work}
\label{sec:related_work}

\subsection{\NER{} for Speech}
\label{ssec:speech_ner}

% \begin{enumerate}
%     \item related work, important - no one goes back to audio annotation
%     \item ETAPE, ESTER - language, small, no audio annotation?
%     \item other datasets?
% \end{enumerate}

Prior work addressed entity recognition for audio recordings via the \emph{audio \NER{}} task: the detection of entities in the text transcript of the audio input.
The majority of these works used a pipeline approach, in which \ASR{} is first applied to the audio and then \NER{} is applied on the noisy textual output of the \ASR{}. These works include discriminative models \cite{sudoh2006incorporating}, incorporating OOV word indicators \cite{parada2011oov}, hierarchical structure \cite{raymond2013robust}, and conditional random fields \cite{hatmi2013named}.

Many audio \NER{} works learn from and measure performance on French datasets, such as \ESTER{} \cite{galliano2009ester} and \ETAPE{} \cite{galibert2014etape}. This may indirectly affect the overall quality of these systems because the \ASR{} component, which is crucial in the pipeline approach but is typically used ``off-the-shelf'', has lower performance in languages other than English.
%In addition, we believe it would be preferable to use an English dataset for evaluation as English is the standard language in medical record datasets (e.g.\ \cite{chiu2017speech}). 

An alternative \ete{} approach was proposed by \citet{ghannay2018end}, in which the model accepts audio as input and outputs a tagged word sequence which consists of normal words and the \NER{} labels in HTML-like tag encoding. Their model did not attain reasonable performance, perhaps due to the small training set.

We emphasize that both pipeline and \ete{} approaches output tagged word sequences, and do not propagate the recognized entity labels back for redaction on the audio itself, which is the end goal of our proposed audio \DEID{} task.

\subsection{De-identification in the Health Domain}
\label{ssec:text_deid}
    
% Main points:
% \begin{enumerate}
%     \item related work
%     \item end: we use lample/dernoncourt because its state of the art
% \end{enumerate}

Previous efforts of \DEID{} in health care focused on redaction of textual medical records. The main approach involves applying \NER{} techniques to the text, including rule-based \cite{ruch2000medical, neamatullah2008automated} and machine learning \cite{guo2006identifying,yang2015automatic} methods. 

Adoption of neural network models boosted the performance of \NER{} on text without requiring hand-crafted rules and complex feature engineering \cite{Collobert2011,Huang2015NNbasedNER,Lample2016NNbasedNER,Ma2016,Dernoncourt2017}. \citet{dernoncourt2017identification} applied the model proposed in \citet{Lample2016NNbasedNER} to medical \DEID{}, achieving state-of-the-art performance on the I2B2-2014 % (denoted \IFOURTEEN{})
\cite{I2B214} \DEID{} challenge dataset.
We have chosen this architecture for the \NER{} component of our pipeline method (Section~\ref{sec:models}).
%and the \MIMIC{} de-identification dataset assembled by the authors, respectively. 

\section{The Audio De-identification Task}
\label{sec:audio_deid_task}

% %Main points:
% \begin{enumerate}
%     \item Goal of Audio Deid, audio in audio out, max recall mi redactions from the conversation
%     \item Formal definition - audio, transcript, identifying, interval, redaction and output audio
%     \item Evaluation metric - TP, FP, FN, coverage ratio
% \end{enumerate}

The goal of the Audio \DEID{} task is to convert an input audio stream into a modified audio stream where the \IDENT{} words are redacted. In essence, the goal of the task is to limit the ability of a listener to identify the entities of the conversation while leaving as much information as possible in order to keep the audio understandable. 

Formally, the input audio stream is a function $\Audio{t}$ of time, that can be transcribed into a sequence of words $W=\{\word{j}\}$, where $\word{j}$ is mapped to the time interval in the audio $T_j = [t_j^{start}, t_j^{end})$. We consider each word to be either \IDENT{} or \NONIDENT{}, and let $I$ denote the set of \IDENT{} words $\{j:w_j \mbox{ is \IDENT{}}\}$.

The output of an audio \DEID{} algorithm is a zero-one redaction function $R(t)$, indicating which parts of the audio stream are to be redacted, where a value of zero indicates \IDENT{} information at time $t$. 
The redacted audio stream can be obtained by zeroing out the redacted part of the stream, $A_{redacted}(t)=R(t)A(t)$.

%This translates to an entity Recall/Precision (or  Sensitivity/Specificity) question.

To evaluate the performance of a \DEID{} algorithm, we term $w_j$ as {\em fully-covered} if $R(t)$ is zero for all $t\in T_j$, and define a corresponding indicator function $\FULLCOVERED{w_i}$. This in turn defines the following standard \NER{} metrics for the audio \DEID{} task:
\begin{eqnarray*}
TruePositives\ (TP)\!\!&=&\!\!\sum_{j \in I} \FULLCOVERED{w_j}, \\
FalsePositives\ (FP)\!\!&=&\!\!\sum_{j \not \in I} \FULLCOVERED{w_j}, \\
%TrueNegatives\ (TN)\!\!&=&\!\!\sum_{j \not \in I} 1-\FULLCOVERED{w_i} \\
FalseNegatives\ (FN)\!\!&=&\!\!\sum_{j \in I} 1-\FULLCOVERED{w_j} \\
Precision=\frac{TP}{TP + FP}\!\!\!\!&\!\!\!\!,\!\!\!\!&\!\!\!\!Recall=\frac{TP}{TP + FN}\\
\end{eqnarray*}
%This metric leads to the standard definitions of Recall/Precision and Sensitivity/Specificity, namely that Sensitivity relates to the number of entity words that are redacted, and Specificity relates to the  number of non-entity words that were redacted.

Finding the exact time interval corresponding to a word is not a trivial task, while redacting most of the interval $T_j$ results in a similar \DEID{} effect as fully covering all the interval.
%, because of text-to-audio alignment errors. 
To this end, we extend $\FULLCOVERED{w_i}$ 
%by introducing a parameter $\rho$ that stands for a threshold percentage out of the time interval of every \IDENT{} that should be redacted by $R(t)$. We then define
into the indicator $\PARTCOVERED{w_j}$ that is true iff $R(t)$ is zero on at least $\rho$ proportion of interval $T_j$. 

With this indicator function we further extend the aforementioned \NER{} metrics to $TP_{\rho}$, $FP_{\rho}$, and $FN_{\rho}$, and correspondingly define $Recall_{\rho}$ and $Precision_{\rho}$. When $\rho=1$ these metrics equal the strict metrics. When $\rho<1$ the new metrics determine the quality of the system with respect to redacting at least $\rho$ of each audio interval in \IDENTS{}.

%Many medical models employ the Sensitivity and Specificity evaluation metrics, as the ratio of true negatives is a good indicator for the effectiveness of the model. In our case, due to the small percentage of entity-related words, the expected specificity was extremely high and non-indicative of the model's performance, therefore we chose to use Recall and Precision for evaluation purposes.

We note that the proposed metrics only evaluate the redaction function $R(t)$ on the word intervals.
% , and not on the gaps between words, which we deem of secondary importance
% \footnote{It is not difficult to amend the definitions above to address also word gaps, should the need arise.}.

%Our audio de-id pipeline maps the input A to audio stream DEID(A) = REDACT(A, T) * F where REDACT(A, T) %is the redaction function of audio and its transcription, resulting in R = Diag({r_i}), r_i \in {0, 1} %where r_i = 0 iff frame f_i should be redacted by giving the frame a zero value.
%The pipeline approach involves first transcribing the audio to text, then running a neural NER %classifier on the transcripts which labels the words w.r.t. the different entity tags, and finally %aligning the recognized words back to the audio timings for redaction. The whole pipeline flow can be %seen in Figure 1.

\section{Datasets}
\label{sec:datasets}

% % Main points:
% \begin{enumerate}
%     \item Datasets - type, statistics, labels and aggregated label groups for comparability
%     \item Datasets - creating training data - preprocessing, labeling
% \end{enumerate}

% Dataset statistics table
\begin{table}
\centering
\scriptsize
\begin{tabular}{ |l|l|r|r|r| } 
 \hline
 \multicolumn{1}{|c|}{\multirow{2}{*}{Dataset}} & 
 \multicolumn{1}{|c|}{\multirow{2}{*}{Medium}} & 
 \multicolumn{1}{|c|}{\multirow{2}{*}{\# Notes}} & 
 \multicolumn{1}{c|}{\multirow{2}{*}{\# Tokens}} & 
 \multicolumn{1}{l|}{\multirow{2}{*}{\% \IDENT{} }} \\
& & & & \\
 \hline
 \IFOURTEEN{} train & Text & 521 & 336,422 & 3.5 \\
 % set1: 11869 PHI tokens/336,422 = 3.5
 \hline
  \CALLISTO{} train & Audio & 4,629 & 8,348,899 & 0.02 \\
%  \hline
%  \SWITCHBOARD{} (train) & \multirow{2}{*}{{Audio}} & 200 & 247,452 & 2 \\
%  \SWITCHBOARD{} (test)  &                          & 50 & 60,957 & 1.7 \\
%  \hline
%  \FISHER{} (train) & \multirow{2}{*}{{Audio}} & 268 & 462,869 & 2.3 \\
%  \FISHER{} (test)  &                          & 58 & 97,966 & 2.5 \\
 \hline
 \SANDF{} train & \multirow{2}{*}{{Audio}} & 468 & 710,348 & 1.8 \\
 \SANDF{} test  &                          & 108 & 158,923 & 2.0 \\
 \hline
\end{tabular}
\caption{Dataset statistics for train and test sets, showing the number of notes (written or spoken), token count, and percent of tokens which are \PHI{}.}
\label{tab:datasets}
\end{table}

% Labels distributions table
\begin{table}
\centering
\scriptsize
\begin{tabular}{ |l|r|r|r| } 
  \hline
  \multicolumn{1}{|c|}{\PHI{} Labels \%} &
  \multicolumn{1}{|c|}{\IFOURTEEN} &
  \multicolumn{1}{|c|}{\CALLISTO{}} &
  \multicolumn{1}{|c|}{\SANDF{}} \\
  & & & train / test \\ % & train / test \\
 \hline
\textbf{Name} & 0.84\% & 0.12\% & 0.22\% / 0.23\% \\ \hline
\textbf{Age} & 0.24\% & 0.01\% & 0.12\% / 0.1\% \\ \hline
\textbf{Date} & 1.56\% & 0.03\% & 0.1\% / 0.12\% \\ \hline
Hospital & 0.28\% & 0.004\% & - \\ \hline
Pharmacy & - & 0.01\% & - \\ \hline
\textbf{Organization} & 0.025\% & 0.003\% & 0.48\% / 0.59\% \\ \hline
\textbf{Location (General)} & 0.001\% & 0.004\% & 0.24\% / 0.29\% \\ \hline
State & 0.07\% & - & 0.15\% / 0.16\% \\ \hline
\textbf{City} & 0.08\% & 0.003\% & 0.25\% / 0.29\% \\ \hline
Country & 0.02\% & - & 0.23\% / 0.27\% \\ \hline
Profession & 0.04\% & - & 0.23\% / 0.27\% \\ \hline
Holiday & - & - & 0.12\% / 0.07\% \\ \hline
Season & - & - & 0.04\% / 0.03\% \\ \hline
\hline
\end{tabular}
\caption{Statistics for \PHI{} labels as percent of total tokens per dataset. Tags in \textbf{bold} are common to all datasets and are used in Section \ref{sec:results}}
\label{tab:labels}
\end{table}

\begin{table}
\centering
\scriptsize
\begin{tabular}{ |l|r|r|r|r| } 
 \hline
  \multicolumn{1}{|c|}{Word} &
  \multicolumn{1}{|c|}{\multirow{2}{*}{\WER{}}} &
  \multicolumn{1}{|c|}{Well} &
  \multicolumn{1}{|c|}{Extended} &
  \multicolumn{1}{|c|}{Shortened} \\ 
 \multicolumn{1}{|c|}{Type} & & Aligned & Alignment & Alignment \\
 \hline
%   \SWITCHBOARD{} & 39 & ?  \\
%  \hline
%   \FISHER{} & 31 & ? \\
%  \hline
  \IDENT{} & 41.8 & 86\% & 5\% & 9\% \\
%  \hline
  \NONIDENT{} & 38.3 & 81\% & 8\% & 12\% \\
 \hline
\end{tabular}
\caption{\ASR{} \WER{} and token-audio alignment distribution on sample conversations from the \SANDF{} dataset.
}
\label{tab:asr-performance}
\end{table}

To create a benchmark for the audio \DEID{} task, we use three datasets from two distinct domains: \emph{conversational English} and \emph{medical records}.
We summarize the main dataset statistics in Table \ref{tab:datasets}.
Importantly, we did not perform text normalization specific to each domain.

In the domain of medical datasets, we use \IFOURTEEN{} \cite{I2B214}, which consists of identified textual medical notes with \IDENT{} tagging, and the Audio Medical Conversations dataset from \cite{chiu2017speech}, denoted \CALLISTO{}, which contains de-identified audio of doctor-patient conversations and their corresponding manual transcripts. 
Processing the \CALLISTO{} conversations was facilitated by the fact that it is a de-identified dataset, which provides us with the locations of the \PHI{} in the audio and the transcripts. Three \PHI{} types: names, dates and ages were redacted, preserving type information, and synthetic data was generated using dictionaries and context-aware rules. First names were drawn from the US Social Security Administration babies names registry\footnote{\url{ https://www.ssa.gov/oact/babynames/}} and last names were drawn from the Frequently Occurring Surnames list from the 1990's US Census\footnote{\url{https://www.census.gov/topics/population/genealogy/data/1990_census.htmlcensus_namefiles.html}}. 
Human annotators used surrounding context to resolve the other \PHI{} types and filled in fake appropriate identifiers.

Notably, neither of the above-mentioned medical datasets could serve as a benchmark for the audio \DEID{} task, as \IFOURTEEN{} is text-based, and \CALLISTO{} contains only redacted audio conversations and is not publicly available. Therefore, we focused on the conversational English domain, where we generated a combined dataset \SANDF{} from the \SWITCHBOARD{} \cite{Godfrey:1992:STS:1895550.1895693} and \FISHER{} \cite{David04thefisher} datasets. 
These datasets include hundreds of conversations in English about a variety of subjects, along with their transcripts. 
To enable proper training and evaluation for the audio \DEID{} task, we annotated all 250 \SWITCHBOARD{} conversations, and 326 from \FISHER{}. Annotation included named \IDENT{} labels, and the time intervals $T_j = [t_j^{(start)}, t_j^{(end)})$ matching each named \IDENT{} back into the audio. This dataset is publicly available$^{\textcolor{darkblue}{\href{ftn:data-public-url}{\ref{ftn:data-public-url}}}}$ to allow for standardized evaluation of novel approaches to this task.

The annotation process began by tokenization of the transcripts provided in both datasets using white-space separators, removing special transcript characters and keeping word capitalization in its original form. Following that, \IDENT{} word annotation was performed manually. The results can be seen in Table \ref{tab:labels}.\\ 

As performing temporal labeling manually is an arduous process, we opt for a semi-automatic \ASR{}-based procedure. 
To this end, we determine word start and end times by aligning the manual transcripts to audio intervals.
We assess the quality of this semi-automatic labeling scheme using human evaluation. 
For a random sample of 6 \SANDF{} conversations (3 \SWITCHBOARD{} and 3 \FISHER{}), we slice the audio according to the aligned interval times per transcript word, and measure both the quality of the transcription, and that of the alignment. 
Table \ref{tab:asr-performance} shows the distribution of alignment errors of the tokens from the sample conversations. These are denoted as good alignment, short (i.e. \ASR{} interval is shorter than actual word) and extended (i.e. interval is longer than expected) where all alignment errors are in the scale of 30-60ms (1-2 audio frames).
% During our \NER{} annotation of the transcripts, we encountered several transcription errors

% Delete comment block if the above is satisfactory

% Evaluation of the audio \DEID{} task requires labeling intervals \intervalj{} $ = [t_j^{(start)}, t_j^{(end)}|$ in the audio with tags, which is an arduous and complicated task, therefore we chose to create semi-automatically rendered labels, and \REPHRASE{verify them with humans}. 
% First, the manual transcripts of both datasets were annotated with a set of fine-grained labels, which were then amalgamated into a coarse set of labels that are shared across datasets, for comparability. 
% The results can be seen in Table~\ref{tab:labels}. The transcripts were tokenized by whitespace, special transcript characters were removed and words were left with their original capitalization.
% To add the temporal labeling, we utilized the \SWITCHBOARD{} and \FISHER{}'s utterance-level temporal transcriptions, which not only include the text of each utterance but also its start and end time in the audio. 
% Using forced-alignment on the utterance transcript and the corresponding interval in the audio were were able to create word-level temporal alignment, and when combined with the aforementioned textual annotations gave us full \ete{} \DEID{} labels.

\section{Pipeline Models}
\label{sec:models}

% Main points:
% \begin{enumerate}
%     \item different models from the datasets and reasoning
%     \item different model parameters - \# hypotheses, ASR alignment vs. forced alignment
%     \item Different models and details of each one, how we trained them and their h-parameters.
%     \item Main conversation model - parameters and experiments they will see - coverage threshold vs. padding, number of hypotheses, TP and FP vs. coverage threshold
% \end{enumerate}

\begin{figure}
    \centering
    \includegraphics[width=0.25\textwidth]{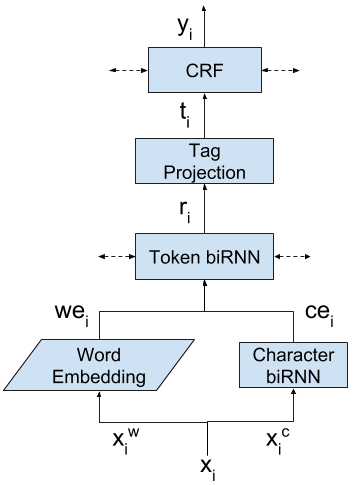}
    \caption{Neural architecture for text \DEID{}}
    \label{fig:text_deid}
\end{figure}

We next describe the models we trained and evaluated to gain insights on the types of challenges this task presents. 
We chose to use the pipeline approach as an audio \DEID{} benchmark due to the ubiquity and maturity of the \ASR{} technology, and abundance of training data for text \NER{}. Our pipeline models contain three main components:
\begin{enumerate}
    \item \label{pipeline-asr} An \ASR{} system, which transcribes the audio into text.
    \item A \NER{} tagger, which tags the transcript with the required labels.
    \item An alignment component, which maps each word in the transcript back to its time interval in the audio. 
\end{enumerate}

For the \ASR{} component, we use Google Cloud's Speech API\footnote{\url{cloud.google.com/speech-to-text}} with the \textit{command\_and\_search} model, which gave us the best transcription accuracy on the data. 
For each conversation, which usually contains two different speakers, we send the entire audio to the service to obtain the transcript. The API also returns alternative hypotheses for the corresponding text and their confidence. We incorporate these alternative hypotheses by taking the top-k \ASR{} hypotheses and feeding them into the next two stages. We then take the logical OR of the detections on all of the hypotheses. 
Unless stated otherwise, $k=1$.

For the \NER{} tagger component, our models use the architecture described in \citet{Lample2016NNbasedNER}, depicted in Fig. \ref{fig:text_deid}. 
This is a neural network model using pre-trained GloVe word embeddings\footnote{\url{nlp.stanford.edu/data/glove.6B.zip}} \cite{Glove} and a character-based bi-directional RNN to generate token embeddings, followed by a bidirectional RNN, tag projection, and CRF layers. 
% The overall architecture is presented in Figure~\ref{fig:text_deid}. 
We define three models, where each model has a \NER{} tagger trained on a different dataset. The models are:
\begin{description}
    \item [\CallistoModel{}] -- Trained using the training data from the \CALLISTO{} dataset.
    \item [\ConversationModel{}] -- Trained using the training data from the \SANDF{} dataset.
    \item [\ItbtModel{}] -- Trained using the training data from the \IFOURTEEN{} dataset.
\end{description}

The \CallistoModel{} and \ConversationModel{} models were trained using the conversation transcripts. We use data augmentation in order to increase robustness to \ASR{} errors, in particular to word deletion, insertion, substitution, and inconsistent capitalization. Data augmentation is carried out in several stages. First we create an \ASR{} transcript from the audio, align it back to the reference transcript by minimizing the word-level edit distance, and transfer the labels to the new transcript. 
For each of the two transcripts, we then generate three additional transcripts by changing word capitalization to camel, lower and upper case. 
Finally, each of the augmented transcripts is broken down into segments of 20 speaker turns with a step of 10 turns, to resemble the utterance structure of the \ASR{} output.
We include the three variants of the \ConversationModel{} model: 
% \begin{enumerate*}
    \ConversationModelOrig{} uses no augmentations,
    \ConversationModelMix{} uses mix-case augmentations only,
    and \ConversationModelMixAsr{} uses all mix-case and \ASR{} augmentations.
% \end{enumerate*}

The \ItbtModel{} model is tuned to achieve state-of-the-art results on textual medical notes, such as in \citet{Dernoncourt2017, Liu2017deid}. It should be stressed that the model was used as is, without an attempt to adapt it to the domain of \ASR{} output. \CallistoModel{} and all \ConversationModel{} models are both trained on conversational data, and should be better adapted to the task. \CallistoModel{} is trained on data originating from the medical domain, as opposed to \ConversationModel{} models which train on data from the English conversation target domain. This is offset by the fact that \CallistoModel{} is trained on a significantly larger training set.

Finally, for the alignment component we add a padding hyperparameter allowing a variable number of mismatched frames at either side of the identified intervals. This slack in interval size is used to compensate for alignment errors.

% Each conversation was tokenized on whitespace/punctuation

% high WER, especially on \IDENT{} words

%The network hyper-parameters used character embedding size of 15, 56 BiLSTM layers, dropout probability 0.4, a batch size of 40, GloVe word embeddings, and 0.2 learning rate. 
% Additionally, we tested our model both with Softmax (ref) and CRF (ref:lafferty) predictive layer, and got better results with CRF. 

%\TODO{In a table} The model's parameters were character embedding size of 25, 64 BiLSTM layers, dropout probability 0.5, a batch size of 90, GLOVE embeddings, and 0.2 learning rate.

\section{Experimental Settings}
\label{sec:experimental_settings}

To test the performance of our models on the audio \DEID{} task, we conducted a number of experiments, described next. 
Section \ref{sec:results} then details our results. We report \RECALL{}, \PRECISION{}, and \FOne{} scores for all experiments, which are significantly more informative than accuracy due to a low \IDENT{}/\NONIDENT{} ratio. We report results on the \SANDF{} test set using the tags which are shown in bold in Table \ref{tab:labels}.
We evaluate our performance against the coverage threshold $\rho \in [0,1]$ which is defined in Section \ref{sec:audio_deid_task}.
Specifically, we focus on type-less metrics, as we care more about the tokens' redaction than their type classification.

Our first experiment evaluates the performance of \CallistoModel{}, \ConversationModel{}, and \ItbtModel{} on the \SANDF{} test set. First, to decouple their tagging performance from the other pipeline errors, we measure their tagging performance on the manually annotated transcripts (referred to as \textit{\NER{} score}). 
\NER{} errors may arise due to train-test disparity, where the train and test data are from different domains or different mediums (e.g.\ text vs. audio), which results in different discriminative models.
Additionally, we measure their overall \ete{} score.
%when they are used in the pipeline
We analyze the complex behavior of the models' precision by inspecting the coverage distribution of \IDENT{} and \NONIDENT{} tokens.

Our second experiment evaluates the effect of two significant hyperparameters on pipeline performance using the \SANDF{} test set:
\\
\begin{itemize}
    \item The number of alternative hypotheses passed on from the \ASR{} to the \NER{} tagger.
    \item The amount of padding added around each detection by the alignment component.
\end{itemize}

%     \item Different models and details of each one, how we trained them and their h-parameters.
%     \item Main conversation model - parameters and experiments they will see - coverage threshold vs. padding, number of hypotheses, TP and FP vs. coverage threshold
% \end{enumerate}

\section{Results} 
\label{sec:results}

% Results were drawn using the colab:
% https://colab.corp.google.com/drive/1ZKMelXqwBaJ13KphqxOvXnqMyWnR1Efx

% % Main points:
% \begin{enumerate}
%     \item \TODO{Data - what labels are we using?}
%     \item All models \begin{enumerate}
%         \item ASR errors per dataset and train/test: WER, PHI WER and Alignment error
%         \item NER and e2e score of each model
%         \item discussion - why we think results are like that,
%     \end{enumerate}
%     \item Conversation model \begin{enumerate}
%         \item Recall and precision w.r.t. padding
%         \item Recall and precision w.r.t. hypotheses
%         \item discussion - Precision behavior w.r.t.\ coverage threshold and why, leading to the FP vs. TP w.r.t.\ coverage threshold
%         \item discussion - ASR + NER + alignment errors are lower than overall error, so we're doing something right
%     \end{enumerate}
% \end{enumerate}

% NER models results
\begin{table}
\centering
\scriptsize
\begin{tabular}{ |l|c|c| } 
 \hline
 \multicolumn{1}{|c|}{\multirow{2}{*}{Model}} & 
 \multicolumn{1}{|c|}{\multirow{2}{*}{\NER{}}} & 
 \multicolumn{1}{|c|}{\multirow{2}{*}{\FOne{} ($\rho$)}} \\ 
 & Recall / Precision / F1 &  \\
 \hline
 \hline
 \ItbtModel{} & 0.37 / 0.48 / 0.41 & 0.37 (0.4) \\
 \hline
 \CallistoModel{} & 0.18 / 0.98 / 0.3 & 0.23 (0.35) \\
 \hline
 \ConversationModelOrig{} & 0.82 / 0.92 / 0.87 & 0.41 (0.4) \\
 \hline
 \ConversationModelMix{} & 0.87 / 0.92 / 0.89 & 0.46 (0.4) \\
 \hline
 \ConversationModelMixAsr{} & 0.88 / 0.92 / 0.9 & 0.51 (0.4) \\
 \hline
\end{tabular}
\caption{\NER{} score of the different \NER{} models, and their \ete{} \FOne{} in their optimal choice of $\rho$.}
\label{tab:ner-models-results}
\end{table}

% Conv model error analysis
\begin{table}
\centering
\scriptsize
\begin{tabular}{ |l|r|r| } 
 \hline
 \multicolumn{1}{|c|}{\multirow{2}{*}{Error type}} & 
 \multicolumn{1}{|c|}{\multirow{2}{*}{Count}} & 
 \multicolumn{1}{|c|}{\multirow{2}{*}{\% of total}} \\ 
 & & \\
 \hline
 \hline
 \ASR{} Transcription errors & 152 & 45.24 \\
 \hline
 \NER{} errors & 168 & 50 \\
 \hline
 Alignment errors & 14 & 4.17 \\
 \hline
 Manual Transcription errors & 2 & 0.6 \\
 \hline
\end{tabular}
\caption{Error analysis of a sample of \ConversationModel{} $FN$ errors, including errors from all components across the pipeline and even occasional manual transcription errors which contribute to both $FP$ and $FN$ errors.
% \TODO{This is not interesting, maybe show how many times we got it right anyway? or that we are better than asr + ner + alignment errors}
}
\label{tab:model-error-analysis}
\end{table}

\begin{figure*}
    \centering
    \includegraphics[width=0.87 \textwidth]{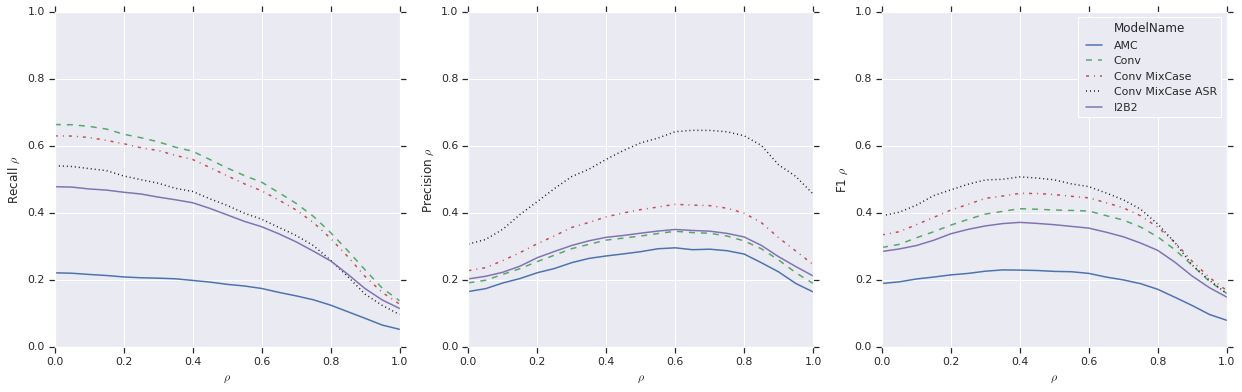}
    \vspace{-.2em}
    \caption{\ETE{} performance comparison of \NER{} models -- Recall (left), Precision (middle) and F1 (right).}
    %\ConversationModel{} variants show a strong advantage.}
    \vspace{2em}
    \label{fig:models-comparison}
    \includegraphics[width=0.87 \textwidth]{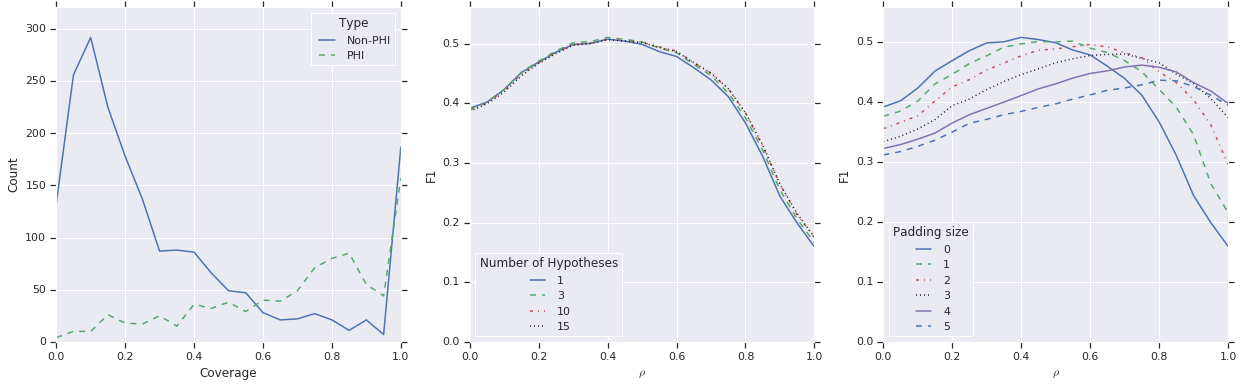}
    \vspace{-.2em}
    \caption{Coverage distribution between \IDENT{} and \NONIDENT{} tokens (left) and
    \ETE{} performance for different pipeline parameters -- number of hypotheses (center) and padding (right). }
    \label{fig:params-comparison}
\end{figure*}

In Table \ref{tab:asr-performance} we report the Word Error Rate (\WER{}) of our \ASR{} component on the \SANDF{} dataset, which was computed by comparing the manual and \ASR{} transcripts of the entire audio. For \WER{} of \IDENT{} words, we removed all the \NONIDENT{} words from manual \ASR{} transcripts before computing the \WER{}. \WER{} of non-PHI words was computed similarly.
We see that both \WER{}s are substantial, and can be thought of as an upper-bound on our pipeline's \ete{} performance.

% \TODO{Why ASR gives lower recall}

Next, Table \ref{tab:ner-models-results} shows the \NER{} and their \ete{} performance of each model for its \FOne{} optimal choice of coverage threshold $\rho$.
We can also see that the \ConversationModel{} surpasses the others in performance due to its training set being in-domain and in the same medium.
Additionally, the \ConversationModelMixAsr{} variant does not display any advantage over its other variants when running on manual transcripts, but gets significantly better performance on the \ete{} scenario.
The difference between \NER{} and \ete{} scores is apparent, and may be attributed to additional pipeline components of \ASR{} and alignment.
Interestingly, in the case of \ConversationModelOrig{}, compounding the \WER{} and alignment error rate from Table \ref{tab:asr-performance} and the \NER{} from Table \ref{tab:ner-models-results} leads to an expected \RECALL{} of approximately 0.44, yet the \ete{} \RECALL{} at $\rho = 0.5$ is 0.53. 
This implies a non-trivial co-dependence between errors in the different components of the pipeline.

Figure \ref{fig:models-comparison} presents the \ete{} evaluation of the different models with respect to the coverage threshold $\rho$.
As expected, $Recall$ is monotonically non-increasing with respect to the threshold. 
Meanwhile, $Precision$ (and consequently $F1$) are not monotonic and have more complex behavior. This behavior is due to difference in the distribution of the coverage between \IDENT{} and \NONIDENT{}, which we see in Figure \ref{fig:params-comparison} (left).
An interesting insight is that most \IDENT{} words have more than half their length redacted by the pipeline while \NONIDENT{} words' coverage is bi-modal, one mode close to 0, and the other close to 1.
% Analysis \TODO{make sure} shows that 
A plausable explanation for this behavior is that the $FP$s are derived from alignment errors in low coverage, while the high coverage $FP$s occur due to classification errors, either due to \ASR{} transcription mistakes or due to model \NER{} errors.

% \begin{figure}
%     \centering
%     \includegraphics[width=0.4 \textwidth]{figures/positives_negatives_coverage.png}
%     \caption{Normalized coverage distribution between \IDENT{} and \NONIDENT{} tokens.
%     Low coverage of the \NONIDENT{} tokens derived from alignment errors, high coverage from \NER{} errors.}
%     \label{fig:positives-negatives-coverage}
% \end{figure}

Finally, we show the \ete{} evaluation of the pipeline using \ConversationModelMixAsr{} with different choices of the pipeline parameters.
In Figure \ref{fig:params-comparison} (center) the performance of the pipeline slightly increases when using additional alternative hypotheses, while a different experiment shows that when using alternative hypotheses with \ConversationModelMix{} performance decreases.
%, which can be attributed to the additional robustness to noise due to training on \ASR{} outputs
% a more significant the increase in recall than drop in precision, which is consistent with 
This decrease is consistent with the hypotheses' decreasing confidence scores, which can be alleviated with \ASR{} training data but is not addressed by the naive OR approach described in Section \ref{sec:models}.
% but rather due to the model's increased robustness to \ASR{} outputs by being trained on augmented data. 
This leads us to seek new ways to utilize the additional \ASR{} artifacts, such as the hypotheses confidence scores and speech lattice.
In Section \ref{sec:conclusions} we discuss possible directions to improve the pipeline's robustness to \ASR{} errors.
Last, Figure \ref{fig:params-comparison} (right) shows that the choice of padding size does not improve performance, but rather alters the value of the optimal coverage threshold.

\section{Conclusions}
\label{sec:conclusions}

% % Main points:
% \begin{enumerate}
%     \item Defined a new task - audio deid, evaluation metrics and a new published dataset
%     \item experiments showed that domain matters, data amount matters, and that most errors come from ASR
%     \item Future work includes alleviating ASR errors: Diarization for better transcripts and alignments, extending multiple hypotheses to the actual lattice (ref lattice RNN). Also, maybe more data on manually labeled audio
% \end{enumerate}

We introduced the audio \DEID{} task, an important prerequisite for protecting privacy when processing sensitive audio datasets in the medical domain as well as other domains. To this end, we created and made available a new test set benchmark derived from annotating the \SWITCHBOARD{} and \FISHER{} audio datasets. We also presented new metrics for the task, \RECALLRHO{} and \PRECISIONRHO{}, as extensions of standard \RECALL{} and \PRECISION{} where words are considered de-identified when at least a portion $\rho$ of their audio signal is redacted.
Finally, we detailed our algorithm for this task, a pipeline approach consisting of three components: \ASR{}, \NER{} on transcripts and a novel alignment from tagged transcripts to audio for the actual redaction. 

We showed that \ASR{} performance is the main impedance towards achieving results comparable to text \DEID{}. 
% Additional safety measures could be employed to increase the system's privacy guarantees, such as rule-based heuristics and biasing the model towards recall, both of which increase the model's \NER{} score's recall at the expense of precision. 
% However, we believe that directly improving the model is the right course for increased privacy guarantees.
In future work, we plan to address this through several directions, including \ete{} \DEID{} \cite{ghannay2018end}, lattice-based techniques \cite{Ladhak+2016}, and diarization and segmentation of the audio as part of the transcription process \cite{cerva2013speaker}.

% \ignore{
% In this paper we defined the audio \DEID{} task, an important prerequisite for protecting privacy and sharing sensitive datasets, which enable application development in medical and other relevant domain. We introduced the evaluation metrics and used them to measure our pipeline's performance as a new benchmark on our newly published \SWITCHBOARD{} and \FISHER{} labels, and publish our temporal labels for these conversations and provide them for public use. 
% We find that correct selection of pipeline hyper-parameters, such as padding and coverage threshold, can lead to performance improvement.
% We show that \ASR{} performance is the main impedance towards achieving results comparable to text \DEID{}, and requires non-trivial integration into the model.
% Additionally, we showed that models trained on in-domain data and mediums have a significant advantage.

% Future work will focus on tackling the \ASR{} errors.
% One future direction is the \ete{} approach, similar to \cite{ghannay2018end}, which will bypass the need to use \ASR{} and may allow the model to implicitly use speech acoustic features for \NER{}, but will require novel methods for creating adequate amounts of labeled data. This could be made feasible using simulation techniques such as \cite{simonnet2018simulating}.
% Another direction would be to leverage lattice-based techniques, such as in \cite{Ladhak+2016}, or to incorporate diarization and segmentation of the audio into the transcription process, which has been shown to improve transcription performance \cite{cerva2013speaker}.
% }

\section{Acknowledgements}
\label{sec:acknowledgements}
The authors would like to thank Oren Gilon, Shlomo Hoory, Amir Feder, Debby Cohen, Amit Markel, and Ronit Slyper for their generous help in the writing of this paper.

Deidentified clinical records used in this research were provided by the i2b2 National Center for Biomedical Computing funded by U54LM008748 and were originally prepared for the Shared Tasks for Challenges in NLP for Clinical Data organized by Dr. Ozlem Uzuner, i2b2 and SUNY.

\bibliography{main.bib}
% \bibliography{main.bbl}
\bibliographystyle{style/acl_natbib}
\end{document}